\documentclass[tikz,pdflatex,sn-mathphys-num]{sn-jnl}
\usepackage{float}

\usepackage[utf8]{inputenc}
\usepackage{graphicx}%
\usepackage{multirow}%
\usepackage{amsmath,amssymb,amsfonts}%
\usepackage{amsthm}%
\usepackage{mathrsfs}%
\usepackage[title]{appendix}%
\usepackage{xcolor}%
\usepackage{textcomp}%
\usepackage{manyfoot}%
\usepackage{booktabs}%

\usepackage{newunicodechar}
\newunicodechar{₉}{$_9$}
\usepackage[utf8]{inputenc}
\usepackage{algorithm}%
\usepackage{algorithmicx}%
\usepackage{algpseudocode}%
\usepackage{listings}%
\usepackage{float}%
\usepackage{multicol} %
\usepackage{academicons} %
\let\orcidlogo\relax
\usepackage{orcidlink}
\usepackage{booktabs}   
\usepackage{makecell}   
\usepackage{amssymb}    
\usepackage{multirow}
\usepackage{booktabs}
\usepackage{graphicx}
\usepackage{tikz}
\usepackage{caption}
\usepackage{amsmath}   
\usepackage{booktabs}  
\usepackage{multirow}  

\theoremstyle{thmstyleone}%
\theoremstyle{thmstyletwo}%

\theoremstyle{thmstylethree}%

\usetikzlibrary{arrows.meta, positioning, calc, fit, backgrounds}

\definecolor{cGray}{HTML}{5F5E5A}
\definecolor{cGrayFill}{HTML}{F1EFE8}
\definecolor{cBlue}{HTML}{185FA5}
\definecolor{cBlueFill}{HTML}{E6F1FB}
\definecolor{cTeal}{HTML}{0F6E56}
\definecolor{cTealFill}{HTML}{E1F5EE}
\definecolor{cAmber}{HTML}{854F0B}
\definecolor{cAmberFill}{HTML}{FAEEDA}
\definecolor{cPink}{HTML}{993556}
\definecolor{cPinkFill}{HTML}{FBEAF0}
\definecolor{cPurple}{HTML}{534AB7}
\definecolor{cPurpleFill}{HTML}{EEEDFE}

\newcommand{\tl}[2]{{\sffamily\small\bfseries #1}\\[1.5pt]{\sffamily\scriptsize #2}}

\begin{document}

\title[Article Title]{Cross-seed explainability using Procrustes conditioned Joint End-to-end Top-K Sparse Autoencoders}


\author[1,2]{Bendegúz Váradi\,\orcidlink{0009-0008-2744-4062}}

\author[1,2]{Zoltán Kmetty\,\orcidlink{0000-0002-6775-8938}}

\affil[1]{\orgdiv{Centre for Social Sciences, CSS-RECENS Research Group, Budapest 1097, Hungary}}

\affil[2]{\orgdiv{Department of Sociology, Faculty of Social Sciences, Eötvös Loránd University, Budapest, Hungary}}


\abstract{
We present a Procrustes-conditioned Joint End-to-end Top-K Sparse Autoencoder (SAE) for extracting cross-seed universal features from independently trained BERT models. Cross-seed feature universality is a fundamental challenge in mechanistic interpretability: because dictionary learning is non-convex, independently trained networks learn misaligned feature spaces, so apparently identical features may differ by random initialization. We address this by computing an orthogonal Procrustes rotation between seeds' activation spaces before joint SAE training, combining Top-K sparsity, end-to-end downstream optimization, and an auxiliary dead-feature revival loss based on previous SAE literature. Evaluating on five independent seed pairs (ten BERT models) across three benchmark datasets (SST-2, Stanford Politeness, TweetEval Emotion), our full pipeline produces more universal features (Pearson \textit{r} $\geq$ 0.70 across seeds) than post-hoc alignment baselines on all three datasets. A minimal qualitative analysis confirms that high-universality features encode interpretable sociolinguistic patterns.

}




\maketitle

\section{Introduction}\label{sec1}

In recent years, a large body of research has been published on the interpretability of language models \cite{shu2025survey}. Much of the recent literature aims to solve two fundamental issues: addressing polysemanticity \cite{cunningham2023sparse, bricken2023towards, gao2024scaling, rajamanoharan2024improving, galichin2026have} and improving the structural fidelity and functional robustness of the extracted features \cite{gao2024scaling, braun2405identifying, marks2024enhancing}.  In this paper, we propose an End-to-End pipeline that combines recent novel methodologies to enhance the reliability of extracting interpretable concepts from BERT models. To demonstrate the feature universalities, we use three benchmark corpora to measure the feature correlations.

Sparse Autoencoders (SAEs) have emerged at the forefront of mechanistic interpretability for disentangling polysemantic representations in large language models \cite{shu2025survey}. Recent advancements have rapidly improved the structural fidelity of these extracted features. TopK SAEs \cite{gao2024scaling} eliminate the shrinkage bias inherent in L1
regularization, End-to-End SAEs \cite{braun2405identifying} prevent feature splitting by optimizing for downstream consistency, and Orthogonal SAEs \cite{anonymous2025orthogonal} (under review at the time of citation) mitigate feature absorption by applying competition-aware orthogonality constraints to disentangle co-activating concepts.
Despite these single-model improvements, achieving cross-seed "universal interpretability" \cite{thasarathan2025universal} remains a fundamental challenge. Because dictionary learning is non-convex, independently trained networks, even those sharing the same architecture and data, suffer from feature splitting, where identical semantic concepts map to entirely different latent dimensions. Recent solutions, such as the Feature Aligned SAE \cite{marks2024enhancing}, attempt to mitigate this by training multiple SAEs in parallel with a Mutual Feature Regularization (MFR) penalty to encourage decoder similarity. However, relying solely on training penalties may fail to address the underlying geometric misalignment of the models' native activation spaces.
In this paper, we present a novel architecture that directly solves this spatial misalignment through a Procrustes-conditioned Joint End-to-End Top-K Sparse Autoencoder. Rather than penalizing separate dictionaries to force alignment, we calculate an Orthogonal Procrustes rotation matrix to superimpose the activation space of different model seeds before extracting concepts with a single, joint SAE. The Top-K activation enforces sparsity structurally, avoiding the L1 shrinkage bias, the Procrustes rotation then provides an optimal orthogonal alignment without the task-loss distortion of previous model-stitching approaches.
We evaluate this method on three English-language benchmark datasets for multi-class classification. Our results demonstrate that this computationally lightweight approach achieves a higher number of cross-seed "universal" features (defined by cross-seed feature Pearson correlation $\geq0.7$) than standard joint-model baselines, providing a promisingly consistent method for extracting linguistic information from black-box models.

\section{Contributions}\label{sec2}

In this paper, we aim to establish a computationally lightweight and consistent method for cross-seed "universal interpretability." Our main contributions are as follows:

1. In order to solve the spatial misalignment problem between independently trained models and to keep the most important previous achievements on SAEs, we introduce the Procrustes-conditioned Joint End-to-End Top-K Sparse Autoencoder. Unlike \cite{anonymous2025orthogonal} who use the orthogonality penalty to disentangle concepts within a single model's latent space, we aim to extract the maximum number of "universal" features from five pairs of models using a joint training architecture following Anthropic's example.

2. We combine the Top-K constraint \cite{gao2024scaling} with a slightly modified version of the end-to-end training objective \cite{braun2405identifying}. By concurrently optimizing for local reconstruction  $\mathcal{L}_{\mathrm{local}}$ alongside downstream mean squared error $\mathcal{L}_{\mathrm{DS}}$ and KL-based output loss $\mathcal{L}_{\mathrm{KL}}$, we push the model toward greater functional consistency without relying on an explicit ${L 
_1}$ sparsity penalty.

3. We apply a pre-computed Procrustes rotation during the training of the joint SAE along with a cross-seed sparse activation loss $\mathcal{L}_{\mathrm{cross}}$ with a similar objective to Marks et al. (2024) \cite{marks2024enhancing}, but instead of using MFR and multiple SAEs, we show that a single SAE is capable of producing highly correlated, "universal" features.

\section{Methodology}\label{sec3}

In recent years the mechanistic interpretability (contrary to the game-theory based Shapley-Additive Values interpretation) of LLMs have been a dynamically moving field of research. Contrary to the SHAP-based method which is computationally demanding and often struggles to generalize among different model initialization seeds (ex. \cite{enouen2025instashap}), the literature on Sparse Autoencoders (SAEs) approaches the issue from a mechanistic perspective. Instead of the input-output pairs, it focuses on mapping the neuron activations into a sparse matrix to reduce feature splitting and encourage mono-semantic interpretability (e.g \cite{lindsey2024sparse}). Numerous versions of SAEs have been developed in recent years. The current study attempts to synthesize some of these approaches. Shu et al. \cite{shu2025survey} present a comprehensive survey on the use of Sparse Autoencoders (SAEs) as a mechanistic interpretability tool to decode the internal representations of Large Language Models (LLMs). One of the most important characteristics of SAEs is the way the they address the concept entanglement of language models by projecting the model's dense representations into a higher-dimensional, overcomplete dictionary where a Top-K sparsity constraint forces each dimension to represent a single, disentangled, and interpretable concept. Shu et al. summarize these recent architectural and training strategy improvements \cite{shu2025survey}.

In our study, we deploy a joint TopK SAE based on \cite{gao2024scaling} to increase the sparsity of the representations which enforces sparsity structurally by retaining only the
\textit{k} largest activations per token. We also experiment with Gao et al.'s Auxiliary loss to reduce the fraction of dead neurons. We integrate \cite{braun2405identifying}'s End-to-end SAE, combining an MSE and a KL Divergence loss calculation to encourage layer-wise consistency.  

\subsection{Joint Top-K End-To-End Sparse Autoencoder with Orthogonal Procrustes alignment}

Because dictionary learning is non-convex, training separate SAEs on different model seeds yields misaligned feature spaces. Several papers have tried to address this issue before, by implementing joint training of encoders \cite{claflin2026feature}, or by "stitching" two networks \cite{csiszarik2021similarity}. Similarly, Anthropic's Joint SAE trains a single SAE for two models encouraging the representational similarity. Getting a consistent latent space between models is crucial in deciding whether the model has learned an interpretable concept or whether the validity of the concept is determined by the random seed, essentially interpreting a model artifact. Comparing learned representations across seeds in this way is closely related to model diffing \cite{claflin2026feature}. Previous studies suggest that universal features can be uncovered by some of these methods outlined below (\cite{lan2024sparse}, \cite{lindsey2024sparse}). While \cite{puri2025atlas} also successfully demonstrate the utility of Orthogonal Procrustes for post-hoc latent alignment, their method relies on a static, pre-trained dictionary that cannot adapt to the cross-seed micro-variations of the subject model.

Csiszárik et al. \cite{csiszarik2021similarity} experiment with Orthogonal Procrustes, where the goal of rotation is to minimize the Frobenius norm. They use this as a similarity metric whereas we use this to align the latent spaces of two models with different seeds. Our paper aims to demonstrate that by combining the aforementioned SAE architecture with an Orthogonal Procrustes rotation of the BERT models' latent spaces, we can achieve a relatively robust cross-seed feature space. 

We let $H_A, H_B \in \mathbb{R}^{n \times d}$ denote the dense activation matrices extracted from the corresponding layers of Model A and Model B for the same sequence of $n$ tokens, where $d$ represents the hidden dimension size ($d = 768$). To map the latent space of Model B into the geometric coordinate space of Model A, we must compute an alignment matrix. The goal is to find an orthogonal transformation $W_{align} \in \mathcal{O}(d)$ that aligns the representations by minimizing the Frobenius norm of their geometric distance:

\begin{equation}
W_{align} = \arg\min_{W \in \mathcal{O}(d)} ||H_B W - H_A||_F
\end{equation}

where $\mathcal{O}(d)$ is the set of $d \times d$ orthogonal matrices.

\subsection{Model Training}

Firstly, a standard BERT model was trained with random initialization and industry standard hyper-parameters (for detailed training parameters see Appendix A).

To increase the validity of the features, we rotate the embedding space as outlined in the previous section. To achieve this, a sample of 500 datapoints were processed by two independent BERT models (justification for the sample size can be found in Appendix B) and the Orthogonal Procrustes rotation matrix was computed on their hidden state $ h_n $ where \textit{n} is the layer id of the model. Later, this rotation matrix was used during the training phase as shown in the training objective.

The training objective is defined as a unified loss:

\begin{equation}
\mathcal{L} = \mathcal{L}_{\mathrm{KL}}
            + \lambda_{\mathrm{DS}}\,\mathcal{L}_{\mathrm{DS}}
            + \mathcal{L}_{\mathrm{local}}
            + \lambda_{\mathrm{cross}}\,\mathcal{L}_{\mathrm{cross}}
            + \lambda_{\mathrm{aux}}\,\mathcal{L}_{\mathrm{aux}}
\label{eq:loss}
\end{equation}

Building on the end-to-end downstream SAE objective of
\cite{braun2405identifying}, which replaces $L_1$ sparsity
regularisation with KL-based output fidelity, we extend the
training objective to joint cross-seed training. Since our
Top-K activation enforces sparsity structurally ($k=32$), no
explicit sparsity penalty is required \cite{gao2024scaling}.
The terms $\mathcal{L}_{\mathrm{KL}}$ and
$\mathcal{L}_{\mathrm{local}}$ are held at unit weight; the
remaining terms carry the tunable coefficients above. Each
component is defined as:
 
\begin{align}
\mathcal{L}_{\mathrm{KL}}    &= \tfrac{1}{2}\bigl[
    D_{\mathrm{KL}}(\mathrm{softmax}(g_A) \,\|\, \mathrm{softmax}(\hat{g}_A))
  + D_{\mathrm{KL}}(\mathrm{softmax}(g_B) \,\|\, \mathrm{softmax}(\hat{g}_B))
  \bigr] \\[2pt]
\mathcal{L}_{\mathrm{DS}}    &= \tfrac{1}{2}\bigl[
    \mathrm{MSE}\bigl(\hat{h}^{(L+1)}_A,\, h^{(L+1)}_A\bigr)
  + \mathrm{MSE}\bigl(\hat{h}^{(L+1)}_B,\, h^{(L+1)}_B\bigr)
  \bigr] \\[2pt]
\mathcal{L}_{\mathrm{local}} &= \tfrac{1}{2}\bigl[
    \mathrm{MSE}_{\mathcal{M}}\bigl(\hat{h}_A,\, h_A\bigr)
  + \mathrm{MSE}_{\mathcal{M}}\bigl(\hat{h}_B,\, W_{\mathrm{align}}\,h_B\bigr)
  \bigr] \\[2pt]
\mathcal{L}_{\mathrm{cross}} &= \mathrm{MSE}_{\mathcal{M}}\bigl(c_A,\, c_B\bigr) \\[2pt]
\mathcal{L}_{\mathrm{aux}}   &= \tfrac{1}{2}\bigl[
    \mathrm{MSE}_{\mathcal{M}}\bigl(\hat{r}_A,\; h_A - \hat{h}_A\bigr)
  + \mathrm{MSE}_{\mathcal{M}}\bigl(\hat{r}_B,\; W_{\mathrm{align}}\,h_B - \hat{h}_B\bigr)
  \bigr]
\end{align}

\noindent where $\mathcal{L}_{\mathrm{KL}}$, $\mathcal{L}_{\mathrm{DS}}$,
and $\mathcal{L}_{\mathrm{local}}$ are averaged over both seeds $A$ and $B$; $\lambda_{\mathrm{DS}}$ is the downstream reconstruction coefficient set to 1.5 and $\lambda_{\mathrm{cross}}$ is the cross reconstruction coefficient set to 1.0. 
$\mathcal{L}_{\mathrm{cross}}$ penalises the direct discrepancy between
their sparse codes.
$g$ and $\hat{g}$ denote clean and SAE-hijacked output logits;
$h^{(L+1)}$ the hidden state of the immediate downstream layer $L{+}1$,
a single-layer simplification of \citet{braun2405identifying};
$\mathcal{M}$ the non-padding token mask;
$W_{\mathrm{align}} = VU^\top$ the orthogonal Procrustes rotation matrix
from $\mathrm{SVD}(H_A^\top H_B) = U\Sigma V^\top$;
and $c_A$, $c_B$ the sparse activations produced by the shared
SAE encoder.

$\lambda_{aux}$ is the auxiliary scaling coefficient (set to $1/32$ in accordance with \cite{gao2024scaling}) and $\hat{r}_{A}$, $\hat{r}_{B}$ denote the auxiliary reconstructions produced by the Top-$k_{aux}$ dead features attempting to estimate the residual errors of the main network. Contrary to Gao et al. we did not use ReLU gating before the forward pass of the auxiliary top-k sparse activations. We note that any downstream effect on feature interpretability is supported only by anecdotal observation, and a more rigorous evaluation would be required to prove it

It is worth noting that $\mathcal{L}_{\mathrm{cross}}$ may seem to violate Goodhart's law by optimizing for cross-seed SAE sparse codes while the feature correlations are measured as a metric of success. We argue that, having separately measured the feature correlations with and without the cross loss, the results show that they are measuring different aspects. Because $\mathcal{L}_{\mathrm{cross}}$ penalizes sparse-code discrepancy directly, the rotation-only ablation ($\lambda_{\mathrm{cross}}=0$) serves as the relevant test of whether cross-seed correlation reflects shared structure rather than the optimization target. Appendix Table D4 further proves this by isolating the effect of the Procrustes rotation.

The SAE class was based on ApolloResearch's original implementation.\footnote{\url{https://github.com/ApolloResearch/e2e_sae/tree/main}}

The k parameter of the autoencoder is set to 32 and the sparse matrix component number is set to 6144.

The encoder and decoder are a simple linear layer with the ReLU activation function replaced by a TopK function. Finally, a normalization is applied to the input hidden state.

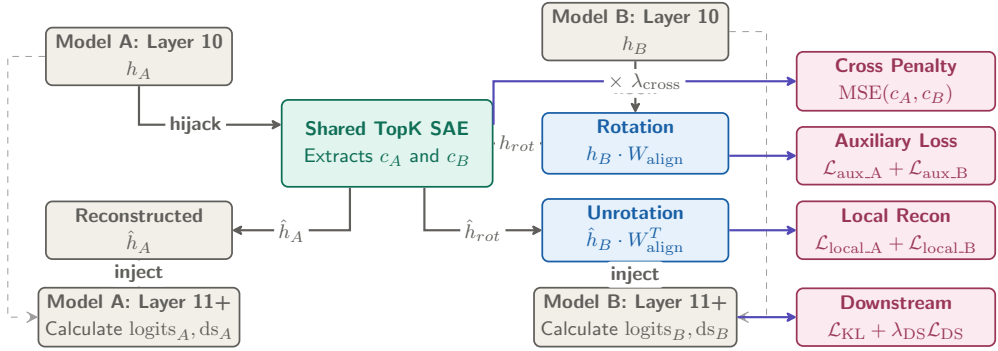
\begin{figure}[htbp]
\centering
\resizebox{\linewidth}{!}{%
\begin{tikzpicture}[
  font=\sffamily\small,
  >={Stealth[round, length=2.0mm, width=2.0mm]},
  base/.style={rounded corners=3pt, draw, line width=0.8pt,
               minimum width=30mm, minimum height=8mm,
               align=center, inner sep=3pt},
  gray/.style ={base, draw=cGray,  fill=cGrayFill,  text=cGray},
  blue/.style ={base, draw=cBlue,  fill=cBlueFill,  text=cBlue},
  teal/.style ={base, draw=cTeal,  fill=cTealFill,  text=cTeal},
  pink/.style ={base, draw=cPink,  fill=cPinkFill,  text=cPink},
  loss/.style ={base, draw=cPink,  fill=cPinkFill,  text=cPink, minimum width=32mm},
  flow/.style={->, line width=1pt, draw=cGray},
  evalflow/.style={->, line width=1pt, draw=cPurple},
  lbl/.style={font=\sffamily\scriptsize\bfseries, text=cGray, fill=white, inner sep=2pt, rounded corners=2pt}
]

\node[gray] (bertA)  at (-4.0,  1.4) {\tl{Model A: Layer 10}{$h_A$}};
\node[gray] (reconA) at (-4.0, -1.4) {\tl{Reconstructed}{$\hat{h}_A$}};
\node[gray] (downA)  at (-4.0, -2.8) {\tl{Model A: Layer 11+}{Calculate $\mathrm{logits}_A, \mathrm{ds}_A$}};

\node[teal, minimum height=14mm, minimum width=34mm] (sae) at (0, 0) 
    {\tl{Shared TopK SAE}{Extracts $c_A$ and $c_B$}};

\node[gray] (bertB)  at (4.0,  1.8) {\tl{Model B: Layer 10}{$h_B$}};
\node[blue] (rotB)   at (4.0,  0.0) {\tl{Rotation}{$h_B \cdot W_{\mathrm{align}}$}};
\node[blue] (unrotB) at (4.0, -1.4) {\tl{Unrotation}{$\hat{h}_B \cdot W_{\mathrm{align}}^T$}};
\node[gray] (downB)  at (4.0, -2.8) {\tl{Model B: Layer 11+}{Calculate $\mathrm{logits}_B, \mathrm{ds}_B$}};

\node[loss] (loss_cross) at (8.2,  1.0) {\tl{Cross Penalty}{$\mathrm{MSE}(c_A, c_B)$}};
\node[loss] (loss_aux)   at (8.2, -0.2) {\tl{Auxiliary Loss}{$\mathcal{L}_{\mathrm{aux\_A}} + \mathcal{L}_{\mathrm{aux\_B}}$}};
\node[loss] (loss_local) at (8.2, -1.4) {\tl{Local Recon}{$\mathcal{L}_{\mathrm{local\_A}} + \mathcal{L}_{\mathrm{local\_B}}$}};
\node[loss] (loss_kl)    at (8.2, -2.8) {\tl{Downstream}{$\mathcal{L}_{\mathrm{KL}} + \lambda_{\mathrm{DS}} \mathcal{L}_{\mathrm{DS}}$}};


\draw[flow] (bertA.south) |- node[lbl, pos=0.7] {hijack} ([yshift=3mm]sae.west);
\draw[flow] ([xshift=-6mm]sae.south) |- node[lbl, pos=0.75] {$\hat{h}_A$} (reconA.east);
\draw[flow] (reconA.south) -- node[lbl] {inject} (downA.north);

\draw[flow] (bertB.south) -- node[lbl] {hook} (rotB.north);
\draw[flow] (rotB.west) -- node[lbl] {$h_{rot}$} (sae.east);
\draw[flow] ([xshift=6mm]sae.south) |- node[lbl, pos=0.75] {$\hat{h}_{rot}$} (unrotB.west);
\draw[flow] (unrotB.south) -- node[lbl] {inject} (downB.north);

\draw[->, dashed, draw=cGray!60] (bertA.west) -- ++(-0.6, 0) |- node[lbl, pos=0.8] {} (downA.west);
\draw[->, dashed, draw=cGray!60] (bertB.east) -- ++(0.6, 0) |- node[lbl, pos=0.8] {} (downB.east);

\draw[evalflow] ([yshift=3mm]sae.east) |- node[lbl, pos=0.75] {$\times\ \lambda_{\mathrm{cross}}$} (loss_cross.west);
\draw[evalflow] ([yshift=-2mm]rotB.east) -- (loss_aux.west);
\draw[evalflow] (unrotB.east) -- (loss_local.west);
\draw[evalflow] (downB.east) -- (loss_kl.west);

\end{tikzpicture}%
}
\caption{Joint SAE Architecture. The shared TopK SAE bottleneck acts as a regularizer, calculating cross, auxiliary, local, and downstream losses simultaneously.}
\label{fig:joint_training}
\end{figure}

During training, each batch is first rotated by the rotation matrix, then passed through the Joint end-to-end TopK SAE implementation. The resulting matrix is transposed to match to original dimensions. Figure ~\ref{fig:joint_training} shows the model training process.

\begin{figure}[h]
\centering
\includegraphics[width=8cm]{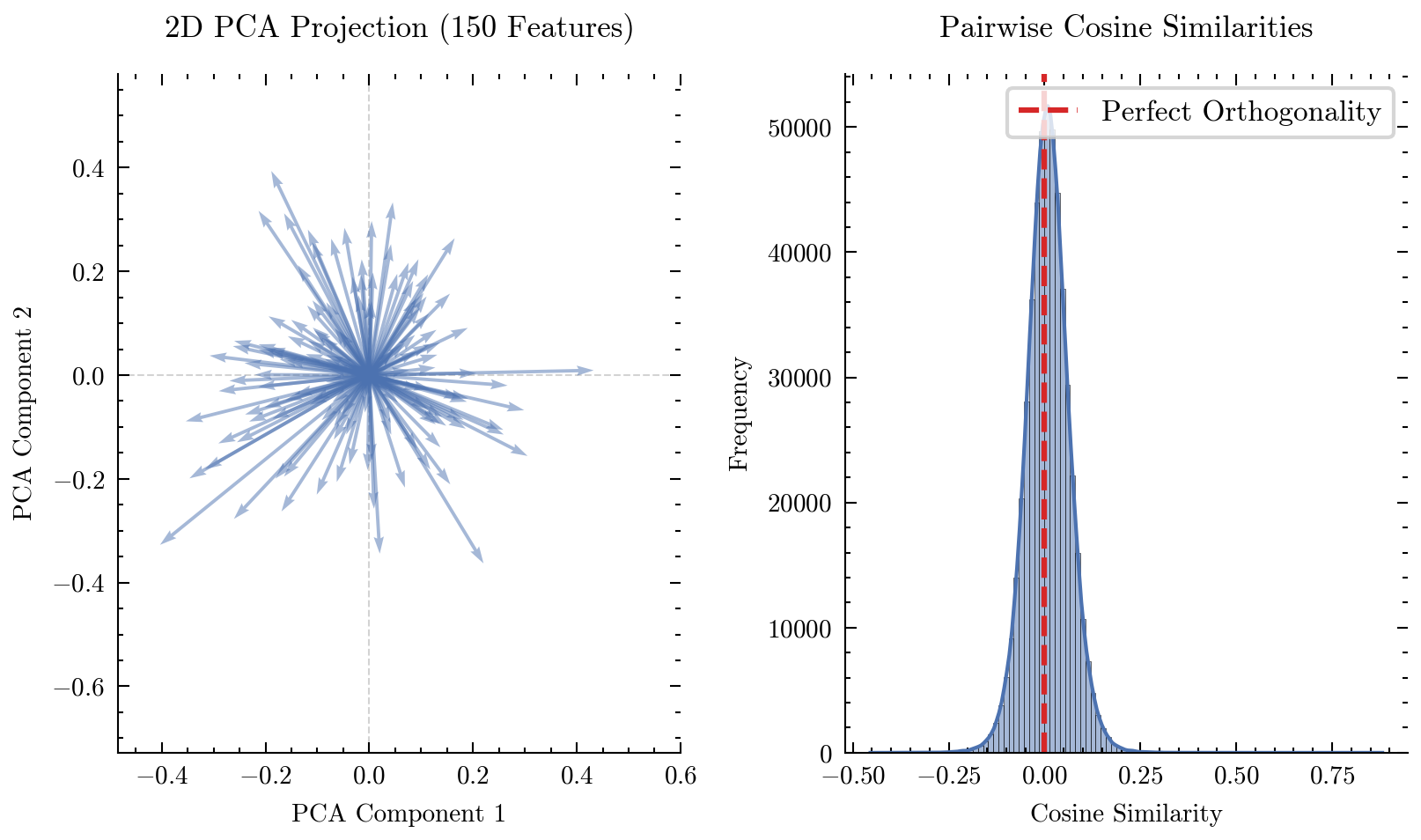}
\caption{PCA projection of SAE features and distribution of SAE component
cosine similarities. The Mean Absolute Cosine Similarity (0.04 in case of the politeness dataset) between features shows
near orthogonality}
\label{fig:pca}
\end{figure}

Figure~\ref{fig:pca}. shows how the Orthogonal Procrustes rotation translates to a 2-dimensional space (PCA). Model A and Model B are independently trained BERT models with different initialisation seeds. The figure shows that the mean absolute cosine similarity between features is relatively low, therefore the semantic entanglement of these features is expected to be lower. 

\section{Evaluation}\label{sec4}

We evaluate the SAEs by comparing the Pearson correlation between matched SAE feature activation vectors across models (top N features by mean activation strength) and datasets. We present the top 10 and top 100 feature correlations this way. Additionally, we note the fraction of dictionary elements that never activate on the evaluation set across either model and the mean classification accuracy change compared to the baseline independent SAE, averaged over Model seed A and Model seed B. We trained the same model architecture on ten different seeds and paired them together for a comparison. We report the mean values in Table 1 along with the standard deviation.

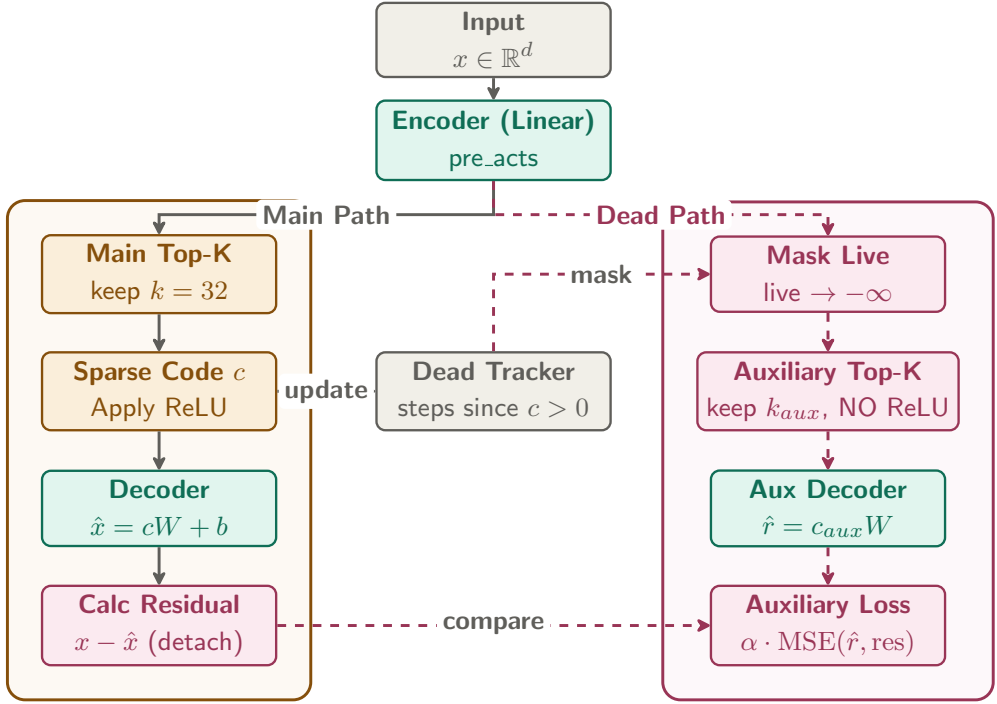
\begin{figure}[htbp]
\centering
\resizebox{\linewidth}{!}{%
\begin{tikzpicture}[
  font=\sffamily\small,
  >={Stealth[round, length=2.0mm, width=2.0mm]},
  base/.style={rounded corners=3pt, draw, line width=0.8pt,
               minimum width=28mm, minimum height=8mm,
               align=center, inner sep=3pt},
  gray/.style ={base, draw=cGray,  fill=cGrayFill,  text=cGray},
  teal/.style ={base, draw=cTeal,  fill=cTealFill,  text=cTeal},
  amber/.style={base, draw=cAmber, fill=cAmberFill, text=cAmber},
  pink/.style ={base, draw=cPink,  fill=cPinkFill,  text=cPink},
  flow/.style={->, line width=1pt, draw=cGray},
  auxflow/.style={->, line width=1pt, draw=cPink, dashed},
  lbl/.style={font=\sffamily\scriptsize\bfseries, text=cGray, fill=white, inner sep=2pt, rounded corners=2pt}
]

\node[gray] (input) at (0,  0.0) {\tl{Input}{$x \in \mathbb{R}^d$}};
\node[teal] (enc)   at (0, -1.2) {\tl{Encoder (Linear)}{pre\_acts}};

\node[amber] (topk)     at (-4.0, -2.8) {\tl{Main Top-K}{keep $k=32$}};
\node[amber] (code)     at (-4.0, -4.2) {\tl{Sparse Code $c$}{Apply ReLU}};
\node[teal]  (dec)      at (-4.0, -5.6) {\tl{Decoder}{$\hat{x} = c W + b$}};
\node[pink]  (residual) at (-4.0, -7.0) {\tl{Calc Residual}{$x - \hat{x}$ (detach)}};

\node[pink] (deadmask) at (4.0, -2.8) {\tl{Mask Live}{live $\to -\infty$}};
\node[pink] (auxk)     at (4.0, -4.2) {\tl{Auxiliary Top-K}{keep $k_{aux}$, NO ReLU}};
\node[teal] (auxdec)   at (4.0, -5.6) {\tl{Aux Decoder}{$\hat{r} = c_{aux} W$}};
\node[pink] (auxloss)  at (4.0, -7.0) {\tl{Auxiliary Loss}{$\alpha \cdot \mathrm{MSE}(\hat{r}, \mathrm{res})$}};

\node[gray] (tracker) at (0, -4.2) {\tl{Dead Tracker}{steps since $c>0$}};

\draw[flow] (input.south) -- (enc.north);
\draw[flow] (enc.south) -- ++(0, -0.4) -| node[lbl, pos=0.25] {Main Path} (topk.north);
\draw[auxflow] (enc.south) -- ++(0, -0.4) -| node[lbl, pos=0.25, text=cPink] {Dead Path} (deadmask.north);

\draw[flow] (topk.south) -- (code.north);
\draw[flow] (code.south) -- (dec.north);
\draw[flow] (dec.south) -- (residual.north);

\draw[flow] (code.east) -- node[lbl] {update} (tracker.west);
\draw[auxflow] (tracker.north) |- node[lbl, pos=0.75] {mask} (deadmask.west);

\draw[auxflow] (deadmask.south) -- (auxk.north);
\draw[auxflow] (auxk.south) -- (auxdec.north);
\draw[auxflow] (auxdec.south) -- (auxloss.north);

\draw[auxflow] (residual.east) -- node[lbl] {compare} (auxloss.west);

\begin{scope}[on background layer]
  \node[rounded corners=6pt, draw=cAmber, line width=1pt, fill=cAmberFill, fill opacity=0.3,
        fit=(topk)(code)(dec)(residual), inner xsep=4mm, inner ysep=4mm] (mainbox) {};
  \node[rounded corners=6pt, draw=cPink, line width=1pt, fill=cPinkFill, fill opacity=0.3,
        fit=(deadmask)(auxk)(auxdec)(auxloss), inner xsep=4mm, inner ysep=4mm] (auxbox) {};
\end{scope}

\node[anchor=north west, text=cAmber, font=\sffamily\small\bfseries] at ([xshift=2mm,yshift=-2mm]mainbox.north west) {};
\node[anchor=north east, text=cPink, font=\sffamily\small\bfseries] at ([xshift=-2mm,yshift=-2mm]auxbox.north east) {};

\end{tikzpicture}%
}
\caption{AuxK Dead Feature Revival Mechanism based on Gao et al. Dead features are trained to reconstruct the detached residual error without ReLU gating.}
\label{fig:auxk_revival}
\end{figure}

\subsection{Datasets}\label{sec5}

For the benchmarking, we choose datasets where the potential features should be fairly well interpretable for the human reader, thus can be qualitatively cross-checked on top of the feature correlations between models.

We choose the sst2 \cite{socher-etal-2013-recursive} dataset for a simple 2-way classification task with simple positive and negative sentiment target classes, the 3-class variant of the Stanford Politeness dataset \cite{danescu2013computational} and the 5-class Twitter emotion classification dataset \cite{mohammad2018semeval}.

\section{Experiments}\label{sec6}

We present the results of the model on three benchmark datasets. The summary of the benchmarks can be found in Table 1. The training parameters for the BERT models as well as the Joint SAE models can be found in the Appendix. We observe six conditions measured on the three datasets. The first condition is reported only to highlight the importance of feature indexing in the embedding space. The fully independently trained SAEs without any rotation or matrix alignment are not expected to yield universal features or meaningful feature correlations since feature indexing is randomly assigned. Therefore, we define the baseline for comparison as the independent SAEs with index matching via a simple Hungarian Algorithm matching on the sparse activations $c_A$ and $c_B$. We do this so that we can separately measure the effect of the Procrustes rotation. The third condition compares independently trained - and subsequently matched - SAEs with Procrustes. The fourth to sixth conditions report the Joint training results while ablating rotation and $\mathcal{L}_{\mathrm{cross}}$; using joint training following Anthropic's pipeline, then joint training with Procrustes rotation but cross-loss disabled ($\mathcal{\lambda}_{\mathrm{cross}}$ = 0); and our full pipeline combining Procrustes conditioning with cross-loss ($\mathcal{\lambda}_{\mathrm{cross}}$ = 1, selected via grid search $\lambda \in \{0.1, 0.5, 1.0, 2.0\}$).

Table 1. shows that compared to independent SAEs, which virtually had zero correlation between features, utilizing a simple feature alignment post-hoc yields a moderately strong correlation between the top 10 and top 100 features. The effect is more pronounced when correlating the top 10 features. The number of active features (univ) with r $\geq$ 0.7 in both models is naturally much higher. Applying Procrustes post-hoc to independent SAEs does not seem to reliably help. It slightly reduces top-10 r on two out of three dataset and changes universal count inconsistently. The rotation's benefit seems to be realised only when paired with joint training, where rotation-only beats joint-no-rotation on every dataset. Joint training in itself however, does not seem to be able to sufficiently direct the models' internal representation towards universality, we can see the most dramatic increase in correlations and universal features in the case of SST-2. 

Regarding cross-loss, we can see a further slight increase in correlations if we also apply the cross-loss set to 1 in all three datasets, but we must note that the improvements are minor compared to the increase in adding Procrustes.

It is also worth noting that even if we look at the top 100 features with the highest activations, we can still see a moderately high correlation between models. The number of dead neurons also increases as we move towards joint training, which is a well-documented phenomenon in TopK SAE training \citep{gao2024scaling}, since the hard 
sparsity constraint concentrates gradient updates on a fixed subset of 
dictionary elements, leaving the remainder without a learning signal. The accuracy, defined by the black box model's classification accuracy injected by the SAE, does not change in a meaningful way.

Table 1 shows the experiments with Auxiliary TopK \cite{gao2024scaling} following Gao et al. who introduced an auxiliary loss on the dead neurons to reactivate them. We can see that this had the strongest effect on the politeness corpus (which has the lowest dead neuron rate), which can be attributed to its average highest sequence length and linguistic complexity among the corpora.

In Table 1. \textit{Top-N r} indicates the mean Pearson $r$ for top-$N$ features by joint mean
activation (active in both models, with at least ${\geq}10$ samples). We define "universality" if a feature active in both models has at least $r \geq 0.70$. We use per-token activations during evaluation (and dead neuron calculation), which is more conservative than sentence-level mean-pooling which can be found in an experiment in the Appendix.

Overall, experiments with Procrustes rotation achieve the best top-10 r, top-100 r, and universal percentage on all three datasets. Table 1. shows the experiments on five seed pairs, with ten different BERT models.

To measure how much of the information was retained by the models after the rotation, we calculate the Frobenius norm of the residual between the rotated source and target hidden states, normalised by the norm of the target. Furthermore, to separate the effect of the joint training from the rotation, we ran a permutation test comparing observed "universal" feature counts against 30 (with 30 rotations the smallest achievable permutation p-value is $1/31 \approx 0.032$) random orthogonal rotations per dataset/pair experiment. The Procrustes rotation exceeded all 30 random baselines in every experiment, and also outperformed the identity and mismatched-correspondence controls. Further results can be found in Appendix \ref{secA4}.

Against the fair post-hoc–matched baseline, the Procrustes-conditioned joint pipeline achieves the highest aggregate cross-seed correlation (top-100 r) on all three datasets and the most universal features on two of three; crucially, applying the rotation post-hoc to independent SAEs yields no consistent benefit, showing that Procrustes alignment is effective specifically when it conditions joint training rather than as a standalone transform.

\begin{table*}[t]
\centering
\caption{%
  Feature universality and correlations across three datasets.
  Univ.\,\%: universal features as a percentage of features active in
  both models (\emph{not} of the full 6{,}144-element dictionary, which is
  deflated by the high dead-feature fraction).
  Dead\,\%: dictionary elements firing on fewer than 10 eval samples
  in either model.
  Acc.\,change: mean accuracy change in percentage points, averaged over both seeds;
  Feature universality is measured at the token level: each
  non-padding token contributes one activation row, and a feature is
  universal if its per-token activation vectors correlate at $r \geq 0.70$
  across models. 
  All values are mean\,$\pm$\,std over five independent seed pairs.
}
\label{tab:ablation}
\resizebox{\textwidth}{!}{%
\begin{tabular}{@{}llrrrrrr@{}}
\toprule
\textbf{Dataset} & \textbf{Condition}
  & \textbf{Top-10 $r$ $\uparrow$}
  & \textbf{Top-100 $r$ $\uparrow$}
  & \textbf{Univ.\,$\uparrow$}
  & \textbf{Univ.\,\%\,$\uparrow$}
  & \textbf{Dead\,\%\,$\downarrow$}
  & \textbf{Acc.\,change\,$\downarrow$} \\
\midrule
\multirow{6}{*}{\textit{Politeness}}
  & Independent SAEs (index)              & $0.009{\pm}0.034$ & $0.001{\pm}0.003$ &     $0{\pm}0$ &  $0.1{\pm}0.2$ & $56.5{\pm}14.1$ & $-0.19{\pm}0.32$\% \\
  & Independent SAEs (matched)            & $0.525{\pm}0.046$ & $0.482{\pm}0.042$ & $236{\pm}124$ & $16.1{\pm}3.3$ & $56.5{\pm}14.1$ & --- \\
  & Independent SAEs + rotation (matched) & $0.532{\pm}0.062$ & $0.462{\pm}0.042$ & $185{\pm}95$  & $12.3{\pm}2.9$ & $\mathbf{51.0{\pm}14.5}$ & --- \\
  & Joint --- no rotation, no cross-loss  & $0.459{\pm}0.096$ & $0.602{\pm}0.050$ & $762{\pm}338$ & $44.9{\pm}2.8$ & $68.1{\pm}14.5$ & $-0.05{\pm}0.37$\% \\
  & Joint --- rotation only               & $0.738{\pm}0.025$ & $0.696{\pm}0.028$ & $777{\pm}377$ & $50.2{\pm}5.3$ & $70.2{\pm}16.2$ & $\mathbf{-0.21{\pm}0.19}$\% \\
  & \textbf{Joint --- full}$^\star$       & $\mathbf{0.764{\pm}0.019}$ & $\mathbf{0.722{\pm}0.026}$ & $\mathbf{889{\pm}493}$ & $\mathbf{57.9{\pm}4.4}$ & $71.1{\pm}16.7$ & $-0.08{\pm}0.22$\% \\
\midrule
\multirow{6}{*}{\textit{SST-2}}
  & Independent SAEs (index)              & $-0.000{\pm}0.017$ & $0.003{\pm}0.007$ &  $0$ &  $0.0$ & $81.9{\pm}3.4$ & $\mathbf{0.00}$\% \\
  & Independent SAEs (matched)            & $0.537{\pm}0.091$ & $0.363{\pm}0.056$ & $23{\pm}12$ &  $5.0{\pm}2.6$ & $81.9{\pm}3.4$ & --- \\
  & Independent SAEs + rotation (matched) & $0.530{\pm}0.101$ & $0.370{\pm}0.043$ & $13{\pm}5$  &  $2.5{\pm}1.3$ & $\mathbf{79.2{\pm}3.8}$ & --- \\
  & Joint --- no rotation, no cross-loss  & $0.156{\pm}0.086$ & $0.220{\pm}0.106$ & $51{\pm}43$ & $11.7{\pm}9.7$ & $88.9{\pm}2.7$ & $\mathbf{0.00}$\% \\
  & Joint --- rotation only               & $0.609{\pm}0.063$ & $0.432{\pm}0.151$ & $55{\pm}41$ & $12.7{\pm}9.1$ & $89.5{\pm}2.2$ & $\mathbf{0.00}$\% \\
  & \textbf{Joint --- full}$^\star$       & $\mathbf{0.702{\pm}0.060}$ & $\mathbf{0.503{\pm}0.135}$ & $\mathbf{78{\pm}54}$ & $\mathbf{19.0{\pm}12.4}$ & $87.5{\pm}5.5$ & $\mathbf{0.00}$\% \\
\midrule
\multirow{6}{*}{\shortstack[l]{\textit{TweetEval}\\\textit{(Emotion)}}}
  & Independent SAEs (index)              & $-0.002{\pm}0.014$ & $-0.000{\pm}0.002$ &    $0$ &  $0.0$ & $73.6{\pm}0.8$ & $-0.06{\pm}0.10$\% \\
  & Independent SAEs (matched)            & $0.646{\pm}0.010$ & $0.511{\pm}0.018$ & $81{\pm}19$ &  $9.8{\pm}2.3$ & $73.6{\pm}0.8$ & --- \\
  & Independent SAEs + rotation (matched) & $0.625{\pm}0.053$ & $0.489{\pm}0.010$ & $67{\pm}18$ &  $8.1{\pm}2.2$ & $\mathbf{65.9{\pm}1.6}$ & --- \\
  & Joint --- no rotation, no cross-loss  & $0.737{\pm}0.026$ & $0.723{\pm}0.033$ & $351{\pm}77$ & $49.3{\pm}8.4$ & $82.8{\pm}0.7$ & $0.06{\pm}0.23$\% \\
  & Joint --- rotation only               & $0.859{\pm}0.015$ & $0.828{\pm}0.015$ & $\mathbf{405{\pm}58}$ & $57.6{\pm}7.5$ & $85.2{\pm}0.7$ & $\mathbf{-0.08{\pm}0.58}$\% \\
  & \textbf{Joint --- full}$^\star$       & $\mathbf{0.865{\pm}0.015}$ & $\mathbf{0.846{\pm}0.016}$ & $398{\pm}54$ & $\mathbf{58.3{\pm}5.7}$ & $85.7{\pm}0.7$ & $0.01{\pm}0.36$\% \\
\bottomrule
\end{tabular}%
}
\end{table*}

\subsection{Minimal counterfactual study}

To demonstrate the validity of the features and to highlight the practical use cases of the methodology, we present an ablation study designed to confirm the extracted features. 

The extracted features can be verified either by an LLM or manually by humans. Table 2. shows an example of this process on a single feature. 

\begin{table}[h]
\centering
\small
\caption{Causal ablation results for Feature 389 (\textit{Hedged Indirect Request}). Each counterfactual applies a minimal edit removing or repositioning the sentence-initial modal verb. Activation strength is reported for the highest-activating token per input. Effect = relative change in peak activation.}
\label{tab:ablation389}
\begin{tabular}{p{4.8cm} p{4.2cm} r r}
\toprule
\textbf{Original} & \textbf{Counterfactual} & \textbf{Edit type} & \textbf{Effect} \\
\midrule
\textit{would} padding [...] be a problem? & \textit{is} padding [...] a problem? & Modal swap & $-$11.8\% \\
\textit{Should} I send you an email tomorrow? & I should send you an email tomorrow. & Mood swap & $-$100\% \\
\textit{would} using a separate dictionary be ok? & using a separate dictionary \textit{would} be ok. & Reposition & $-$24.2\% \\
\textit{Should} we just agree to disagree? & We're not going to agree. & Mood swap & $-$100\% \\
\textit{Would} you mind if I had a question? & I have a question. & Directness swap & $-$100\% \\
\bottomrule
\end{tabular}
\end{table}

First, each top activating token was gathered from a SAE injected BERT model along with the context they appear in. The black box model and SAE trained in the politeness dataset had feature 389 as one of its highest activating features extracted from examples passed through the model from the test set. The top activating tokens are "would", "should", "is", "does", "take" appearing in contexts similar to ones presented in Table 2. Then, minimal grammatical and syntactical changes can be made to the input sequence and the change in the attention activation strength recorded. This way, the researcher can test the activation patterns of the model and get a better understanding of the given feature's meaning for human interpretation. Feature 389 encodes sentence-initial modal verbs in interrogative position, as confirmed by the repositioning ablation. Replacing the modality (changing \textit{would} to \textit{is}) has a negative effect on the activations, while changing the interrogative mood to declarative mood forces the model to default to the [CLS] token and the activation drops to zero. Similarly, repositioning the query word or changing the directness of the sentence shows the effect. 

The above example is just one method to qualitatively confirm the concepts behind the model's features. We presented this simple method to show that mechanistic interpretability can be used for meaningful linguistic analyses and that the Procrustes rotated features are orthogonal enough to be semantically disentangled.

\section{Limitations and discussion}

We believe that the outlined method shows promise for a more robust cross-seed explainability framework, however in this section we would like to outline some of the limitations of the method employed. Firstly, we have only compared models from a single BERT model family. To demonstrate the robustness our combined SAE model, we would need to test this method on several more seeds and flavours of language models. 

We are also aware that, as many have highlighted before (e.g. \cite{lindsey2024sparse}), the interpretation of these features is not always trivial or even possible. Depending on which layer we use for the SAE encoding, the results are well documented to vary greatly (for a detailed analysis, see \cite{dalvi2022discovering}). Our current analysis, we use a late representational layer (layer 10 of 12), where theoretically the correlations are the lowest since the model's dictionary learning diverges over the training process, but we could have used an earlier layer which would have resulted in more shallow grammatical, more general syntactic terms or orthographical features with higher correlations. Furthermore, since we also used end-to-end loss, this layer was the last possible one where it makes sense to use. 

Currently we have only compared the pipeline to post-hoc matched SAEs, but a direct comparison with Feature Aligned SAEs \cite{marks2024enhancing} and Orthogonal SAES \cite{anonymous2025orthogonal} is left for an important future work.

\section{Conclusion}\label{sec13}

In this paper, we present a method to extract and explain "universal" features from independently trained black box models using Procrustes conditioned Joint End-to-end Top-K
Sparse Autoencoders. The main contribution of the paper is the combination of the recent developments on SAEs along with the utilization of Orthogonal Procrustes as a feature alignment method.  We evaluate our results on three benchmark datasets and show that highly correlated ("universal") features can be extracted from independent black box models.

\section*{Declarations}

\subsection{Funding}

Supported by the EKÖP-25 university research scholarship program of the ministry for culture and innovation from the source of the national research, development and innovation fund.
The data collection was funded by the Eötvös Loránd Research Network within the framework of Flagship Research Projects: KÖ–32/2021. The work of Zoltán Kmetty was supported by the Digital Political Footprint project, funded by NKFIH, with a grant number [K-147329]; Nemzeti Kutatási Fejlesztési és Innovációs Hivatal [K-147329].

\subsection{Competing Interests}

The authors have no competing interests to declare that are relevant to the content of this article.

\subsection{Declaration of use of AI}

We declare that we have used Gemini 3.1 Pro and Claude Sonnet 4.6 for coding purposes including script writing and refactoring. Furthermore, the models were used for proof-reading the mathematical formulas defined above. All codes provided by the aforementioned LLMs have been manually proof-read and modified where necessary.

\subsection{Reproducibility}

To better ensure reproducibility, during the training of the BERT and SAE models, we set seeds for Numpy \cite{harris2020array} and Pytorch \cite{paszke2017automatic} to 42. We set \textit{torch.backends.cudnn.deterministic} to True. We used a single NVIDIA GeForce RTX 3090 to train the models with CUDA version 12.8. The training parameters for both BERT and SAE models are reported in the Appendix. We used mixed-precision training.

\bigskip

\begin{appendices}

\section{Model Training}\label{secA1}

Table A1 shows the SAE and BERT traning parameters.

\begin{table}[htbp]
\centering
\small
\caption{Hyperparameters used across all experiments.}
\label{tab:hyperparams}
\begin{tabular}{@{}ll@{\hspace{20pt}}ll@{}}
\toprule
\textbf{Hyperparameter} & \textbf{Value} & \textbf{Hyperparameter} & \textbf{Value} \\
\midrule
\multicolumn{2}{@{}l}{\textit{Base model}} & \multicolumn{2}{@{}l}{\textit{Training}} \\
Model                & \texttt{bert-base-uncased} & Optimiser & AdamW \\
Parameters           & 110M                       & Learning rate & $5e-5$ \\
BERT layer (SAE)     & 10                         & lr scheduler type & cosine \\
\multicolumn{2}{@{}l}{\textit{Fine-tuning and pairing}} & Weight decay & 0.001 \\
Seed pair A \& B     & 42 \& 137                  & Early stopping patience & 5 \\
Seed pair C \& D     & 99 \& 7                    & Gradient accumulation steps & 1 \\
Seed pair E \& F     & 11 \& 110                  & Batch size & 16 \\
Seed pair G \& H     & 25 \& 225                  & Epochs & 5 \\
Seed pair I \& J     & 35 \& 446                  & Warmup ratio &  0.1 \\
\multicolumn{2}{@{}l}{\textit{SAE architecture}} & $k_{aux}$ &  512 \\
Dict.\ size $n_\text{dict}$ & 6{,}144           & $\lambda_\text{DS}$ & 1.5 \\
Sparsity (TopK $k$)  & 32                         & $\lambda_\text{cross}$ & 1.0 \\
SAE parameters       & 9{,}444{,}096              & $\lambda_\text{aux}$ & 1/32 \\
Decoder init.        & Orthogonal                 & Max seq.\ length & 128 \\
Injection layer      & 10                         & \multicolumn{2}{@{}l}{\textit{Procrustes}} \\
SAE inference seed   & 42                         & Alignment samples & 500 \\
Pooling strategy     & Per-token                  & \multicolumn{2}{@{}l}{\textit{Universality filter}} \\
                     &                            & Activity threshold & ${\geq}10$ eval samples \\
                     &                            & Univ.\ threshold & Pearson $r \geq 0.70$ \\
\bottomrule
\end{tabular}
\end{table}

\section{Rotational matrix sample size validation}\label{secA2}

\begin{table}[ht]
\centering
\caption{Validation of the Procrustes alignment sample size.}
\label{tab:procrustes_sample_size}
\setlength{\tabcolsep}{4.5pt}
\begin{tabular}{lccccc}
\toprule
Dataset
  & \makecell{500 seqs \\ $=$ tokens}
  & \makecell{Rot.\ dist. \\ at 500 seqs}
  & \makecell{NRE \\ plateau}
  & \makecell{Cross-split \\ std}
  & \makecell{$k_{95}$ / $k_{99}$} \\
\midrule
SST-2
  & 5{,}478
  & 7.42 \scriptsize{(13.4\%)}
  & \checkmark
  & 0.013
  & 2 / 3 \\
TweetEval
  & 11{,}420
  & 2.54 \scriptsize{(4.6\%)}
  & \checkmark
  & 0.011
  & 3 / 7 \\
Politeness
  & 15{,}267
  & 3.36 \scriptsize{(6.1\%)}
  & \checkmark
  & 0.010
  & 2 / 22 \\
\bottomrule
\end{tabular}
\end{table}

Rot.\ dist.\ is the mean Frobenius
distance to the full-data reference rotation $\mathrm{W}_{\mathrm{ref}}$ at the
paper's exact token count; parenthesised values express this as a percentage of
the theoretical maximum $2\sqrt{d} \approx 55.4$ for $d{=}768$. NRE plateau indicates whether the normalised residual error
curve has flattened at the paper's token count. Top-10 $r$ is from Table~1,
full pipeline.

Cross-split std shows the standard deviation of distances of $\mathrm{W_align}$ rotation matrices across 30 random 50/50 splits.

$\mathrm{k_{95}}$ and $\mathrm{k_{99}}$ show the number of singular components required to explain 95 and 99 percent of the total cross-seed shared structure of $\mathrm{H_{A}^\top H_{B}}$.
  
The post-hoc convergence analysis confirms that the 500-sequence Procrustes
budget used in the paper is sufficient across all three datasets. In terms of
token counts, all three datasets place the paper's operating point at or past the
NRE plateau, with cross-split stability standard deviations of $\leq 0.011$ confirming
that the rotation is reproducible. The extremely low effective rank of $\mathrm{H_A^\top H_B}$
further confirms that the two BERT seeds share a
highly stable representational geometry, which is a prerequisite for the
cross-seed interpretability approach proposed in this paper.

\section{SAE results with mean pooling activation and AuxK}\label{secA3}

In the Table C3 we present the SAE feature correlation results with using the Auxiliary loss and mean pooling activation. We theorized that using ReLU might be too restrictive for the current TopK setup and experimented with removing ReLU and using sentence-level mean pooling instead at the evaluation step. Notably, the results show much more universal features and an lower dead neuron rate, but this can risk interpretability and introduce noise. The reason for the dramatic increase is explained by how mean pooling activation makes a feature fire if it is present anywhere in the sequence, whereas the per-token measure enforces a stricter rule.


\begin{table*}[t]
\centering
\caption{%
  Feature universality and correlations across three datasets.
  \textbf{Top-$N$ $r$}: mean Pearson $r$ for top-$N$ features by joint mean
  activation (active in both models, ${\geq}10$ samples).
  \textbf{Univ.}: features active in both models with $r \geq 0.70$.
  \textbf{Univ.\,\%}: universal features as a percentage of features active in
  both models (\emph{not} of the full 6{,}144-element dictionary, which is
  deflated by the high dead-feature fraction).
  \textbf{Dead\,\%}: dictionary elements firing on fewer than 10 eval samples
  in either model.
  \textbf{Acc.\,change}: mean accuracy change in percentage points
  (clean $\to$ SAE-hijacked), averaged over both seeds; negative = improvement.
  Independent baselines are reported under two pairings: \emph{index} (raw
  dictionary index, near-zero by construction) and \emph{matched} (optimal
  one-to-one assignment). The two post-hoc rows reuse the independent SAEs, so
  their reconstruction (and hence Acc.\,change) is identical to the index row
  (marked ``---'').
  All values are mean\,$\pm$\,std over two independent seed pairs.
  $\star$ = best per column per dataset.
}
\label{tab:ablation}
\resizebox{\textwidth}{!}{%
\begin{tabular}{@{}llrrrrrr@{}}
\toprule
\textbf{Dataset} & \textbf{Condition}
  & \textbf{Top-10 $r$ $\uparrow$}
  & \textbf{Top-100 $r$ $\uparrow$}
  & \textbf{Univ.\,$\uparrow$}
  & \textbf{Univ.\,\%\,$\uparrow$}
  & \textbf{Dead\,\%\,$\downarrow$}
  & \textbf{Acc.\,change\,$\downarrow$} \\
\midrule
\multirow{6}{*}{\textit{Politeness}}
  & Independent SAEs (index)              & $0.011{\pm}0.003$ & $0.002{\pm}0.009$ &     $1{\pm}1$ &  $0.1{\pm}0.1$ & $26.1{\pm}5.1$ & $\mathbf{-0.39{\pm}0.08}$\% \\
  & Independent SAEs (matched)            & $0.629{\pm}0.008$ & $0.589{\pm}0.020$ & $532{\pm}42$  & $26.6{\pm}2.2$ & $26.1{\pm}5.1$ & --- \\
  & Independent SAEs + rotation (matched) & $0.607{\pm}0.002$ & $0.565{\pm}0.029$ & $399{\pm}32$  & $20.0{\pm}1.6$ & $\mathbf{22.1{\pm}4.0}$ & --- \\
  & Joint --- no rotation, no cross-loss  & $0.506{\pm}0.107$ & $0.592{\pm}0.048$ & $1691{\pm}255$ & $50.0{\pm}3.6$ & $36.8{\pm}3.6$ & $-0.35{\pm}0.33$\% \\
  & Joint --- rotation only               & $0.770{\pm}0.030$ & $0.724{\pm}0.021$ & $1786{\pm}452$ & $53.0{\pm}2.7$ & $38.0{\pm}11.7$ & $-0.38{\pm}0.08$\% \\
  & \textbf{Joint --- full}$^\star$       & $\mathbf{0.784{\pm}0.011}$ & $\mathbf{0.750{\pm}0.016}$ & $\mathbf{2038{\pm}495}$ & $\mathbf{59.7{\pm}3.5}$ & $37.5{\pm}11.6$ & $-0.10{\pm}0.04$\% \\
\midrule
\multirow{6}{*}{\textit{SST-2}}
  & Independent SAEs (index)              & $-0.056{\pm}0.036$ & $-0.056{\pm}0.036$ &    $0$ &  $0.0$ & $93.4{\pm}0.2$ & $\mathbf{0.00}$\% \\
  & Independent SAEs (matched)            & $0.691{\pm}0.045$ & $0.418{\pm}0.019$ & $12{\pm}2$   &  $6.8{\pm}1.5$ & $93.4{\pm}0.2$ & --- \\
  & Independent SAEs + rotation (matched) & $0.683{\pm}0.014$ & $0.436{\pm}0.004$ & $14{\pm}4$   &  $6.8{\pm}1.8$ & $\mathbf{93.2{\pm}0.1}$ & --- \\
  & Joint --- no rotation, no cross-loss  & $0.317{\pm}0.052$ & $0.276{\pm}0.008$ &  $2{\pm}1$   &  $1.1{\pm}0.6$ & $94.5{\pm}0.3$ & $\mathbf{0.00}$\% \\
  & Joint --- rotation only               & $0.668{\pm}0.083$ & $0.474{\pm}0.012$ & $12{\pm}2$   &  $6.7{\pm}1.5$ & $95.2{\pm}0.5$ & $\mathbf{0.00}$\% \\
  & \textbf{Joint --- full}$^\star$       & $\mathbf{0.746{\pm}0.010}$ & $\mathbf{0.539{\pm}0.021}$ & $\mathbf{30{\pm}7}$ & $\mathbf{16.8{\pm}5.2}$ & $95.7{\pm}0.6$ & $\mathbf{0.00}$\% \\
\midrule
\multirow{6}{*}{\shortstack[l]{\textit{TweetEval}\\\textit{(Emotion)}}}
  & Independent SAEs (index)              & $0.001{\pm}0.011$ & $0.001{\pm}0.001$ &    $0$ &  $0.0$ & $75.2{\pm}1.4$ & $\mathbf{-0.09{\pm}0.02}$\% \\
  & Independent SAEs (matched)            & $0.762{\pm}0.025$ & $0.587{\pm}0.002$ & $114{\pm}18$ & $15.0{\pm}1.7$ & $75.2{\pm}1.4$ & --- \\
  & Independent SAEs + rotation (matched) & $0.762{\pm}0.019$ & $0.585{\pm}0.012$ &  $98{\pm}12$ & $12.9{\pm}1.0$ & $\mathbf{68.2{\pm}1.8}$ & --- \\
  & Joint --- no rotation, no cross-loss  & $0.816{\pm}0.014$ & $0.782{\pm}0.005$ & $372{\pm}62$ & $54.5{\pm}3.0$ & $83.8{\pm}1.0$ & $0.12{\pm}0.20$\% \\
  & Joint --- rotation only               & $0.899{\pm}0.008$ & $0.880{\pm}0.005$ & $\mathbf{432{\pm}36}$ & $65.5{\pm}1.8$ & $86.4{\pm}0.9$ & $0.47{\pm}0.52$\% \\
  & \textbf{Joint --- full}$^\star$       & $\mathbf{0.915{\pm}0.008}$ & $\mathbf{0.887{\pm}0.004}$ & $418{\pm}31$ & $\mathbf{66.5{\pm}0.5}$ & $86.9{\pm}0.9$ & $0.39{\pm}0.10$\% \\
\bottomrule
\end{tabular}%
}
\end{table*}

\section{Measuring the effect of Procrustes rotation}\label{secA4}

This appendix provides more detailed results of the permutation null test.
We compare the universal feature yield of True Procrustes alignment compared to the unrotated baseline (identity), mismatched control rotations and random rotations. The True rotation strictly dominates the unrotated Identity baseline. We compute the universal yield just like in Table 1. Then we compute a Monte Carlo permutation p-value as the rank of the true count among the random rotations. 

Furthermore, the true rotation produced more universal features than all 30 random orthogonal rotations across every experiment (Monte Carlo empirical $p \approx 0.032$). Crucially, the structurally identical Mismatched matrix, which was computed over shuffled token correspondences, fails to achieve comparable universality. We report the observed feature counts directly. Note that the Univ. counts differ from Table 1. because the table below was run on a subset of data (35 batches per experiment) with cross coef. set to 0, varying only the rotational matrix.

We compute the normalised residual error ($\mathrm{NRE} = \lVert H_B W_{\mathrm{align}} - H_A \rVert_F / \lVert H_A \rVert_F$) between the rotated source and target hidden states as a proxy for alignment quality and regress it on the universal feature count. Alignment quality explains 70.4\% of the variance in universal-feature count (slope=0.839, $R^2$ = 0.704, $p < 10^{-67}$).

\begin{table}[h]
\centering
\caption{Effect of Procrustes rotation. Each experiment was computed with
cross-coefficient set to 0, varying only the rotation matrix.
\textbf{Max (Random)} is the highest universal count among the 30 random
rotations.}
\label{tab:permutation_test}
\footnotesize
\setlength{\tabcolsep}{4pt}
\begin{tabular}{@{}llcccc@{}}
\toprule
\textbf{Dataset} & \textbf{Pair} & \textbf{Identity} & \textbf{True} & \textbf{Mismatched} & \textbf{Max (Random)} \\
\midrule
\multirow{2}{*}{Politeness} & 42--137 & 113 & 212 & 0 & 1 \\
                            & 99--7   & 218 & 324 & 0 & 1 \\
\midrule
\multirow{2}{*}{SST-2}      & 42--137 & 17  & 79  & 2 & 0 \\
                            & 99--7   & 18  & 84  & 2 & 2 \\
\midrule
\multirow{2}{*}{TweetEval}  & 42--137 & 287 & 417 & 0 & 1 \\
                            & 99--7   & 237 & 364 & 0 & 1 \\
\bottomrule
\end{tabular}
\end{table}

\begin{figure}[H]
\centering
\includegraphics[scale=0.8]{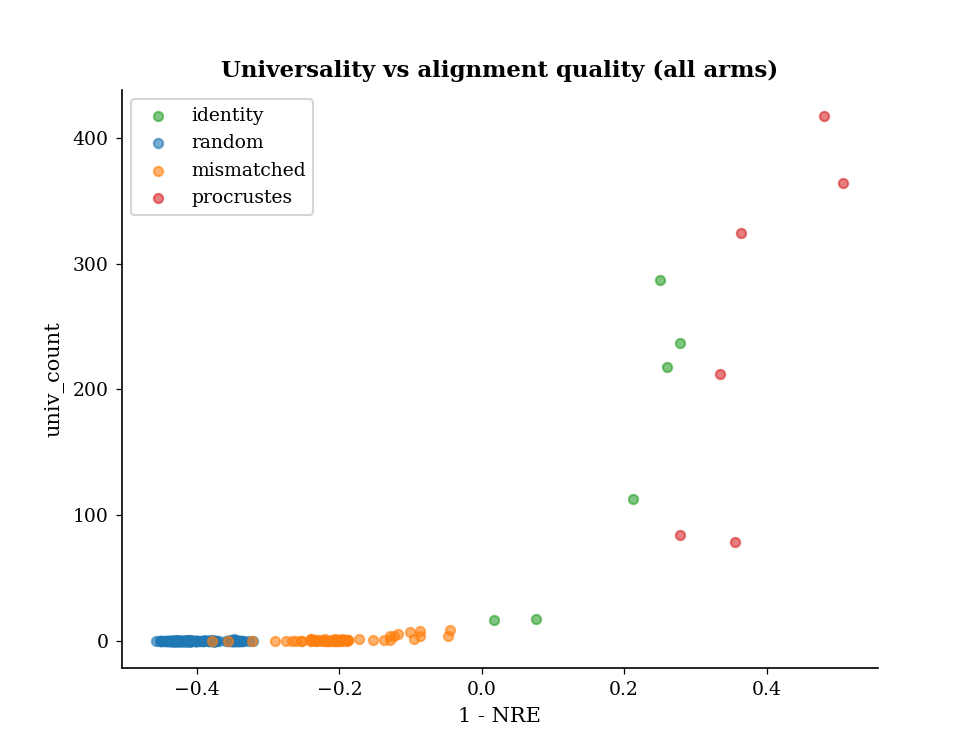}
\caption{Universal-feature yield versus alignment quality (1 - NRE) across all rotation conditions and experiments.}
\label{fig:alignment_quality}
\end{figure}





\bigskip

\end{appendices}
\backmatter

\bigskip
\clearpage
\bibliography{sn-bibliography}

\end{document}